# The SMART+ Framework for AI Systems

**Laxmiraju Kandikatla, MPharm, CQA**
MaxisIT Inc., Edison, NJ, United States
Aula Fellowship for AI, Montreal, Canada
laxmiraju.kandikatla@gmail.com
ORCID: 0009-0009-8824-6446

**Branislav Radeljić, PhD, SFHEA**
Aula Fellowship for AI, Montreal, Canada
branislav@theaulafellowship.org
ORCID: 0000-0002-0497-3470

**Abstract**
Artificial Intelligence (AI) systems are now an integral part of multiple industries. In clinical research, AI supports automated adverse event detection in clinical trials, patient eligibility screening for protocol enrollment, and data quality validation. Beyond healthcare, AI is transforming finance through real-time fraud detection, automated loan risk assessment, and algorithmic decision-making. Similarly, in manufacturing, AI enables predictive maintenance to reduce equipment downtime, enhances quality control through computer-vision inspection, and optimizes production workflows using real-time operational data. While these technologies enhance operational efficiency, they introduce new challenges regarding safety, accountability, and regulatory compliance. To address these concerns, we introduce the SMART+ Framework—a structured model built on the pillars of Safety, Monitoring, Accountability, Reliability, and Transparency, and further enhanced with Privacy & Security, Data Governance, Fairness & Bias, and Guardrails. SMART+ offers a practical, comprehensive approach to evaluating and governing AI systems across industries. This framework aligns with evolving mechanisms and regulatory guidance to integrate operational safeguards, oversight procedures, and strengthened privacy and governance controls. SMART+ demonstrates risk mitigation, trust-building, and compliance readiness. By enabling responsible AI adoption and ensuring auditability, SMART+ provides a robust foundation for effective AI governance in clinical research.

1. Introduction

AI-driven tools and AI Systems hold significant promise across industries—enhancing diagnostics and clinical trial optimization in healthcare, detecting fraudulent transactions and automating loan risk assessments in finance, and enabling predictive maintenance and real-time workflow optimization in manufacturing. However, these technologies also carry substantial risks if not carefully designed, validated, and monitored (Bouderhem, 2024; Chustecki, 2024; De Micco et al., 2025; Ferrara et al., 2024; Khan et al., 2025; Murdoch, 2021; Panteli et al., 2025). A high-profile example is the Epic Sepsis Model, a proprietary algorithm deployed in numerous hospitals, which failed to identify 67% of patients with sepsis while generating alerts for only 18% of admissions (Habib et al., 2021; Wong et al., 2021). This poor real-world performance far below clinician judgment highlights the potential consequences of relying on AI systems without rigorous oversight. Similarly, imaging AI models trained predominantly on light-skinned patients have demonstrated worse performance for lesions on darker skin tones, leading to underdiagnosis in non-white populations (Cross et al., 2024; Rezk et al., 2022). These cases illustrate how AI systems that perform well in controlled environments can fail in practice, potentially exacerbating healthcare disparities and compromising patient safety.

While no single set of AI ethics principles has been universally agreed, let alone accepted and implemented, several trustworthy AI frameworks offer complementary guidance. The NIST AI Risk Management Framework (2023) provides a risk-based approach to AI system development, emphasizing trustworthiness across the AI lifecycle. The OECD AI Principles (2019/2024) promote innovative and responsible AI that respects human rights and democratic values. The GAO AI Accountability Framework (2021) outlines practical governance, data, performance, and monitoring practices for federal AI systems. Similarly, the EU Ethics Guidelines for Trustworthy AI identify a range of requirements that AI systems should satisfy (EU, 2019). While each framework emphasizes critical aspects of trustworthy AI, none fully addresses the unique needs of clinical research or healthcare deployment.

Building on these insights, the SMART+ Framework introduces a streamlined taxonomy tailored for healthcare AI. It integrates the core principles of Safe, Monitored, Accountable, Reliable, and Transparent (SMART), augmented with Privacy & Security, Data Governance, Fairness & Bias, and Guardrails. By doing so, SMART+ provides actionable guidance to ensure that AI systems across industries operate ethically, reliably, and safely. In the following sections, we review the relevant AI frameworks, detail each SMART+ component, and demonstrating its utility in promoting trustworthy AI adoption. We aim to provide a practical framework for achieving trustworthiness in AI systems by ensuring that all essential parameters (SMART+) are thoroughly addressed before the system is released for real-world use. Beyond deployment, our framework emphasizes the importance of continuous monitoring to maintain performance, safety, and ethical integrity over time. This holistic approach enables stakeholders to apply well-defined principles for evaluating AI systems and demonstrating their trustworthiness throughout the entire lifecycle.

2. Theoretical and Methodological Considerations

The global ecosystem of AI ethics and governance is grounded in multiple international frameworks that collectively define the principles of trustworthy, safe, and human-centered AI. Among the most influential is the EU Ethics Guidelines for Trustworthy AI, which articulate seven core requirements—Human Agency and Oversight, Technical Robustness and Safety, Privacy and Data Governance, Transparency, Diversity and Fairness, Societal Well-being, and Accountability (EU, 2019). These are underpinned by four ethical principles: Respect for Human Autonomy, Prevention of Harm, Fairness, and Explicability. Together, they emphasize that AI must enhance rather than diminish human rights, with strong mechanisms for oversight, data integrity, explainability, and non-discrimination. Complementing the EU's perspective, the NIST AI Risk Management Framework (2023) adopts a risk-based, lifecycle-oriented approach structured around four core functions—Govern, Map, Measure, and Manage. It identifies key characteristics of trustworthy AI systems, including validity, reliability, safety, security, accountability, transparency, and fairness. Similarly, the OECD AI Principles (2019/2024) and GAO AI Accountability Framework (2021) promote transparency, human rights, and ongoing performance monitoring. The GAO's four pillars—Governance, Data, Performance, and Monitoring—offer practical measures for organizations to maintain accountability and reliability throughout the AI lifecycle. The EU Artificial Intelligence Act operationalizes these ethical foundations into enforceable regulatory obligations by categorizing AI systems into unacceptable, high, and limited risk tiers (EU, 2024). High-risk systems—common in healthcare, finance, and public sectors—must comply with specific articles (Articles 8–15), addressing risk management, data governance, technical documentation, transparency, human oversight, and cybersecurity. This ensures that AI systems are safe, traceable, and aligned with human welfare.

The UNESCO's Recommendation on the Ethics of AI expands the ethical scope globally, endorsed by 194 member states. It advances principles such as Proportionality and Do No Harm, Fairness, Sustainability, Privacy and Data Protection, Human Oversight, Transparency, Accountability, and Literacy (UNESCO, 2021). This framework situates AI ethics within a human rights context, promoting adaptive governance and multistakeholder collaboration. In the domain of healthcare, the WHO's Ethics and Governance of AI for Health (2021) focuses on six key principles: protecting human autonomy, ensuring transparency and explainability, fostering accountability, promoting equity, and ensuring sustainability (WHO, 2021). These principles emphasize that AI in health must augment human expertise and uphold patient safety. Meanwhile, the IEEE's Ethically Aligned Design and its P7000 Standards emphasize Human Rights, Well-being, Accountability, Transparency, and Misuse Awareness, translating ethical imperatives into design and engineering practices (IEEE, 2022). Finally, the standard ISO/IEC 42001 introduces the Artificial Intelligence Management System (AIMS)—a certifiable framework that helps organizations implement responsible AI through governance, bias mitigation, risk assessment, and continual improvement (ISO, 2023a).

Collectively, these frameworks converge on common pillars—fairness, privacy, safety, accountability, transparency, and human oversight—serving as the foundation for emerging integrative approaches such as the SMART+ Framework. SMART+ consolidates these global principles into actionable pillars (safety, monitoring, accountability, reliability, transparency, complemented with considerations of privacy & security, data governance, fairness & bias, and guardrails) that can be embedded across the AI system lifecycle as per ISO 42001 for compliant, trustworthy, and human-centric AI.

We propose a practical, evidence-driven framework for establishing and demonstrating trustworthiness in AI systems by integrating lifecycle phases, SMART+ principles, and risk-based governance (Figure 1). The framework recognizes that trust in AI cannot be achieved through isolated checks; rather, it must be systematically embedded across the entire AI lifecycle. Accordingly, we consider all major phases—objective setting, requirements and specifications, design and development, verification and validation, deployment, and operation and maintenance—as interdependent stages where trustworthiness must be intentionally cultivated. Within each phase, we incorporate the SMART+ principles as a unifying structure to ensure that expectations are clearly defined, measurable, actionable, and aligned with ethical and technical quality standards.

A central component of the framework is the integration of risk assessment to account for system complexity, context of use, and potential harm (Kandikatla & Radeljić, 2025). The outcome of this assessment informs the proportional application of SMART+ principles: higher-risk AI systems warrant augmented controls, enhanced validation rigor, and more robust oversight, whereas lower-risk systems require correspondingly lighter governance. This risk-stratified approach ensures that resources are directed where they are most needed while maintaining adherence to principles of responsible innovation. By combining lifecycle orientation, principle-based design, and risk-proportionate governance, the proposed framework offers a holistic and traceable method for evaluating trustworthiness. Moreover, it supports continuous monitoring after deployment to ensure that AI systems remain safe, reliable, and aligned with stakeholder expectations throughout their operational lifespan.

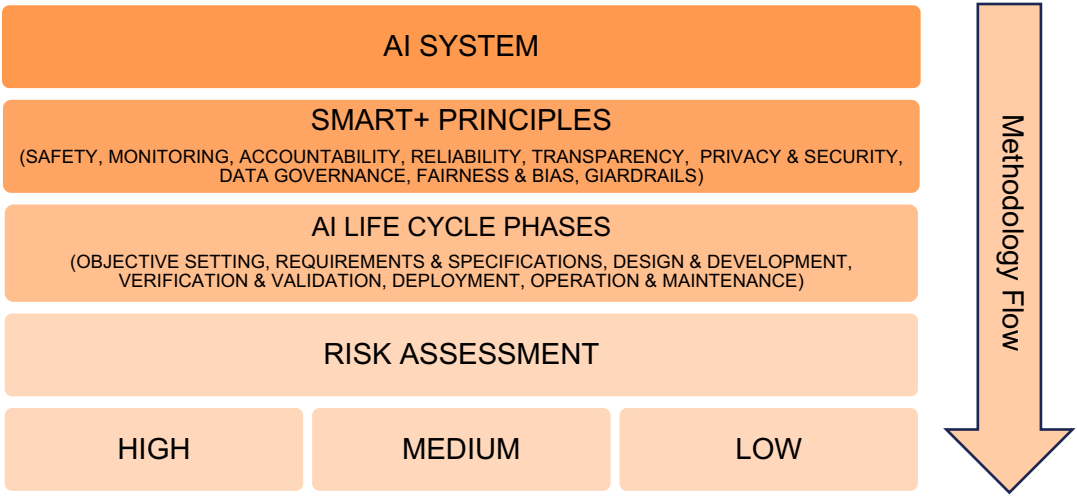

**Figure 1: SMART+ Framework**

## 3. The SMART+ Framework

The SMART+ Framework is a structured model designed to ensure the development, deployment, and monitoring of trustworthy AI systems in regulated and high-stakes domains, such as finance, manufacturing, healthcare, clinical research, and pharmaceuticals. It extends traditional AI governance approaches by integrating multiple dimensions of system quality, accountability, and ethical compliance, each representing a critical aspect of AI trustworthiness (Figure 2).

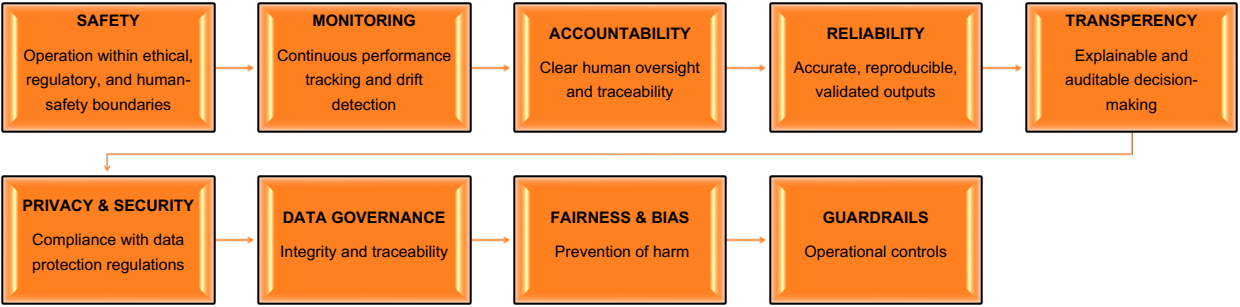

**Figure 2. SMART+ Principles**

By explicitly defining standards and checkpoints under each pillar, SMART+ provides a comprehensive mechanism to identify, mitigate, and monitor risks across the AI lifecycle, aligning with global guidelines. Its structured approach allows organizations to systematically assess AI systems, ensuring operational safety, data integrity, ethical compliance, and regulatory readiness, while also providing measurable criteria for audit and governance purposes. In other words, SMART+ can be systematically embedded into the AI lifecycle. It translates ISO 42001's principles—they represent a foundational shift in organizational AI governance, providing a cohesive and risk-based management system that aligns responsible innovation with compliance, accountability, and ethical stewardship across the AI lifecycle—into a set of operational dimensions that shape both practice and assurance in contemporary AI management (Figure 3).

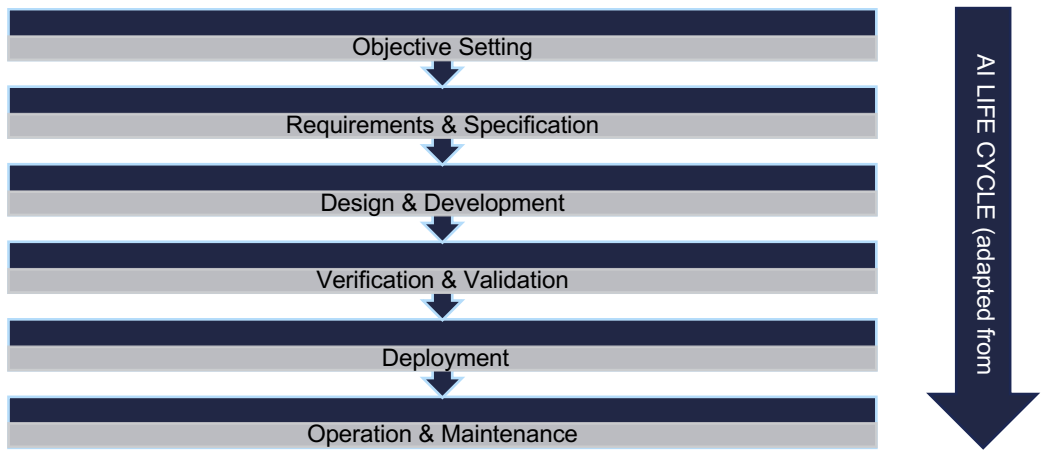

**Figure 3. AI System Lifecycle Management**

The onset of the AI lifecycle begins with rigorous objective setting, where organizations are required to articulate the scope, intended purpose, and explicit ethical boundaries for AI systems. This stage necessitates alignment with contextual stakeholder needs and expresses the organization's commitment to both internal governance and external regulatory expectations. By applying SMART+, safety concerns are proactively assessed to prevent undue risk, while lines of accountability are clearly demarcated, assigning specific ownership for decision-making and oversight. Transparency is established through thorough documentation of the rationale, intended uses, and anticipated impacts of each system. Supporting this, robust data governance practices and clearly defined guardrails ensure that all data usage, behavioral criteria, and policy constraints are both measurable and auditable from inception.

As the lifecycle progresses to the requirements and specifications phase, the operationalization of SMART+ ensures that all technical, functional, and compliance-related expectations are delineated in measurable terms. Data protection is addressed through privacy- and security-by-design principles that map to regulatory imperatives such as the General Data Protection Regulation (GDPR) and the Health Insurance Portability and Accountability Act (HIPAA), or local statutes. The integrity of data—its lineage, quality, and labeling standards—is governed by policies that are explicitly traceable to both internal requirements and societal values. Fairness and bias are proactively managed through systematic assessment, mitigation, and continuous monitoring to ensure that AI outcomes remain equitable across demographic and clinical subgroups. Accountability and transparency are reinforced by demanding requirement traceability not only to technical risks, but also to overarching ethical commitments. Through this, the requirements phase becomes the conduit by which both performance benchmarks and the ethical, societal expectations of AI are captured and operationalized.

The design and development phase is where the system's technical and ethical robustness is engineered. Organizations need to embed controls to ensure the reliability of model architecture, the reproducibility of training processes, and the veracity of data engineering practices. Within SMART+, safety is maintained through systematic identification and mitigation of potential harms, including the management of bias in data and outcomes. Transparency is advanced by requiring comprehensive documentation of design assumptions, the approaches taken to achieve model explainability, and the traceability of data sources. Privacy and security are engineered into every aspect of model development, complemented by access controls and pipeline security. Where risk concentrations are identified, human oversight—whether through 'human-in-the-loop' or 'human-on-the-loop' frameworks (Chiodo et al., 2025; Crootof et al., 2025; Emami et al., 2024; Enarsson et al., 2022; Kandikatla & Radeljić, 2025; Langer et al., 2025; Salloch & Eriksen, 2024; Sterz et al., 2024)—serves as a fundamental guardrail, strengthening operational resilience and ethical assurance.

Verification and validation stand as critical assurance mechanisms, ensuring that the AI system faithfully fulfills its intended use under simulating realistic, real-world conditions. By integrating SMART+ in this phase, organizations undertake rigorous and structured testing for robustness, bias, and sensitivity to data drift or anomalous behavior. Continuous monitoring mechanisms are embedded not only during initial testing, but also as an enduring feature of system operation. Safety validation processes are tailored to address known ethical and functional failure modes, while all phases of validation are meticulously recorded, supporting full accountability and providing a robust evidentiary trail for internal and regulatory audits. Disclosures regarding model limitations and residual uncertainties form part of a culture of transparency and proactive risk disclosure.

Deployment is governed by the principles of controlled release. In this phase, SMART+ assures that deployment is restricted to fully verified systems, with clearly established approval processes and formal oversight. Stakeholder engagement and communication regarding system capabilities, limitations, and intended purposes are not merely procedural, but represent an extension of organizational transparency. Data and process security remain paramount, leveraging technical controls such as encryption and identity management to safeguard both the system and user privacy. Dynamic guardrails, including risk thresholds and automated alerting, are activated before system 'go-live,' ensuring that only ethically aligned and technically sound models reach production environments.

Finally, during the operation and maintenance phase, continual improvement is operationalized through ongoing monitoring, retraining, and audit cycles, ensuring each deployed AI system remains effective, auditable, and ethically aligned in a changing external environment. SMART+ guides the implementation of live performance tracking, drift detection, and systematic anomaly management, reinforcing operational reliability and early incident detection. Defined escalation protocols ensure that issues are addressed within accountable governance structures. Reliability is reinforced through periodic robustness checks and realignment with organizational goals, while transparent reporting mechanisms foster stakeholder trust and facilitate meaningful external scrutiny. In parallel, access controls and data protection measures underpin ongoing privacy and security, while adaptive guardrails govern retraining, system updates, and version control—transforming operational management into a dynamic feedback loop centered on safety, ethics, and trustworthiness.

Consequently, the union of SMART+ with the AI lifecycle standard creates a harmonized and dynamically adaptive governance architecture, capable of sustaining safe, accountable, and ethically robust AI systems throughout the entire lifecycle. By codifying each domain of SMART+ as an operational dimension directly mapped to AI lifecycle requirements, organizations achieve a substantive, multi-dimensional assurance model that simultaneously advances compliance, risk mitigation, ethical accountability, and operational excellence in contemporary AI governance.

## 4. Practical Approach

### 4.1. Safety

The Safety pillar represents the foundational ethical and technical commitment to ensure that AI systems operate without causing harm to individuals, society, or the environment. Safety in AI encompasses the prevention of unintended consequences, robustness against adversarial manipulation, and reliability under varying operational conditions (Choudhury & Asan, 2020; Ferrara et al., 2024; Kandikatla & Radeljić, 2025). As defined by the EU Ethics Guidelines for Trustworthy AI (EU, 2019) and the NIST AI Risk Management Framework (2023), safety entails both technical robustness (resilience, accuracy, and reliability) and ethical soundness (avoidance of harm, bias, or misuse). In high-stakes domains such as healthcare and regulated industries, safety aligns with the "do no harm" principle and the ISO 31000 and ISO 23894 approach to risk management (ISO, 2018 & ISO, 2023)—where potential failure mode is anticipated, quantified, and controlled.

Regarding its integration into the AI lifecycle, Safety must be embedded as a continuous assurance activity throughout all phases:
- *Objective Setting*, through the definition of the AI system's purpose, ethical boundaries, and potential risk domains, as well as the identification of critical safety use cases where harm may occur (e.g., patient misdiagnosis, erroneous recommendations);
- *Requirements & Specifications*, through the establishment of explicit safety requirements and risk tolerance thresholds, as well as the specifications of fail-safe mechanisms, fallback procedures, and performance validation criteria;
- *Design & Development*, through the incorporation of safety-by-design principles, redundancy mechanisms, and guardrail configurations during model training and testing, as well as the implementation of bias detection, adversarial robustness testing, and safety validation checkpoints;
- *Verification & Validation*, through stress testing, scenario simulation, and independent validation to confirm safe performance under edge cases, as well as the verification of conformance to regulatory and ethical safety standards (e.g., ISPE GAMP 5 (ISPE, 2022), ISO/IEC 5338 (ISO, 2023b));
- *Deployment*, through the implementation of human oversight mechanisms for critical decision points, as well as the documentation of incident response plans and establishment of rollback procedures; and
- *Operation & Maintenance*, through the continuous monitoring of AI behavior, performance drifts, and incident reports, as well as periodic safety audits and integration of lessons learned into retraining cycles.

### 4.2. Monitoring

The Monitoring pillar ensures that AI systems are continuously observed for performance consistency, ethical compliance, and operational safety throughout their lifecycle. Monitoring establishes a feedback mechanism for detecting drift, bias, security anomalies, or system degradation in real time (Aljohani, 2023; Corrêa et al., 2023; Khurram et al., 2025; Papagiannidis et al., 2025; Radanliev, 2025). According to the NIST AI Risk Management Framework, continuous monitoring supports the "manage" function, which aims to detect and respond to changes that could undermine trustworthiness. Likewise, ISPE GAMP 5 emphasizes ongoing performance verification, change control, and periodic review of computerized systems. In the context of AI, monitoring extends beyond technical performance to include ethical performance—tracking unintended consequences, data shifts, and user feedback that may affect fairness, safety, and accountability. Therefore, monitoring serves as the dynamic quality control loop that sustains trustworthy AI during both development and operational use. By continuously testing AI (e.g., via shadow mode or simulated data) and incorporating user feedback, organizations catch issues early. This aligns with EU/OECD calls for feedback mechanisms and human oversight loops.

Regarding its integration into the AI lifecycle, Monitoring is applied to multiple phases, ensuring proactive detection and mitigation of deviations from expected behavior:
- *Objective Setting*, through the definition of measurable key performance indicators (KPIs), alert thresholds, and monitoring objectives aligned with risk categories (e.g., high-risk AI systems per EU AI Act Article 9);
- *Requirements & Specifications*, through the inclusion of monitoring and audit requirements (e.g., system logging, data drift detection, explainability audits) in the system specifications;
- *Design & Development*, through the integration of built-in monitoring functions such as model monitoring dashboards, bias detection pipelines, and safety triggers during model evaluation;
- *Verification & Validation*, through the validation of monitoring functions by conducting stress and robustness tests under controlled failure conditions (simulated anomalies or out-of-range inputs);
- *Deployment*, through the introduction of monitoring dashboards, alert mechanisms, and automated logging systems, as well as the definition of escalation paths and responsible personnel for anomaly management; and
- *Operation & Maintenance*, through the performance of continuous monitoring for model drift, accuracy degradation, and ethical compliance, as well as periodic system audits, retraining reviews, and governance board oversight meetings to assess ongoing trustworthiness.

### 4.3. Accountability

The Accountability pillar embodies the principle that responsibility for AI outcomes ultimately lies with humans and organizations, not the autonomous system itself. Accountability ensures that every AI decision, model modification, and system action is traceable, auditable, and attributable to defined human roles (Binns, 2022; Schuett et al., 2025; Solove & Matsumi, 2024; Wagner, 2019). As emphasized by the EU AI Act (Articles 14 & 15), ISO/IEC 42001, and the NIST AI Risk Management Framework, accountability requires clear governance structures, transparent documentation, and mechanisms for redress in case of failure or harm. In regulated environments, accountability aligns with GxP (good practice) principles where each activity—design, testing, deployment, or monitoring—must be verified, approved, and justified by a qualified authority. Thus, accountability operationalizes trust through governance, transparency, and continuous oversight across the AI lifecycle.

Regarding its integration into the AI lifecycle, Accountability should be established and reinforced at each phase through well-defined roles, decision checkpoints, and documentation controls:
- *Objective Setting*, through the definition of governance ownership, including AI sponsors, validation leads, and data stewards, as well as the establishment of a RACI (Responsible–Accountable–Consulted–Informed) matrix to delineate accountability boundaries;
- *Requirements & Specifications*, through the documentation of data provenance, consent management, and intended use, as well as the sign-off on ethical compliance and system purpose statements;

- *Design & Development*, through the maintenance of detailed design history files and version-controlled artifacts that capture model changes, rationale, and developer sign-offs;
- *Verification & Validation*, through the enforcement of independent quality review, traceability matrices, and deviation logs, whereby validation reports must be authorized by accountable personnel (Quality Assurance or Validation Lead);
- *Deployment*, through the approval for release that should follow a structured Change Control and Release Authorization process under the Quality Management System (QMS), and include rollback and escalation procedures; and
- *Operation & Maintenance*, through the maintenance of post-deployment audit trails, incident management logs, and retraining documentation, as well as periodic reviews of accountability structures as part of governance meetings or AI Ethics Board reviews.

### 4.4. Reliability

The Reliability pillar ensures that an AI system performs consistently, accurately, and predictably across different environments, data sources, and use conditions. Reliability in AI reflects the system's ability to maintain intended functionality over time while minimizing unexpected outcomes or failures (Floridi et al., 2018; Pandit & Rintamäki, 2024; Schmider et al., 2019). According to the NIST AI Risk Management Framework, reliability is a core attribute of trustworthy AI, emphasizing robustness, reproducibility, and resistance to drift or degradation. Similarly, ISO/IEC 42001 highlight reliability as a validation outcome, where systems are proven fit for intended use and remain stable throughout their operational lifecycle. In essence, a reliable AI system demonstrates technical robustness, performance stability, and repeatability under real-world and stress conditions, ensuring consistent value delivery without compromising safety or ethics. Overall, Reliability thus acts as the stability assurance bridge between AI model development and its safe, sustained operation.

Regarding its integration into the AI lifecycle, Reliability applies across all phases, with particular emphasis on Design & Development, Verification & Validation, and Operation & Maintenance:
- *Objective Setting*, through the definition of performance expectations and reliability targets such as model accuracy, latency, reproducibility, and robustness criteria, as well as the alignment of reliability metrics with business and regulatory goals (e.g., EU AI Act Article 15);
- *Requirements & Specifications*, through the specification of measurable performance criteria, validation conditions, and test environments in system and data requirements documents;
- *Design & Development*, through the incorporation of reliability engineering principles by building multiple models, testing alternative algorithms, and ensuring reproducibility across datasets and environments, as well as the conduct of robustness tests to evaluate performance under edge cases, missing data, and adversarial conditions;
- *Verification & Validation*, through the performance of formal validation and stress testing to confirm consistent results, as well as the use of statistical verification, confidence intervals, and benchmark comparisons to verify reproducibility, and the implementation of cross-validation and independent testing across temporal or demographic datasets;
- *Deployment*, through the validation of model reliability in real-world conditions via shadow deployment, pilot trials, or A/B testing before full rollout; and
- *Operation & Maintenance*, through the continuous monitoring of key performance metrics and data drift to identify reliability degradation, as well as the implementation of version control and retraining protocols to maintain performance over time.

### 4.5. Transparency

The Transparency pillar ensures that the design, operation, and decision-making processes of an AI system are understandable, explainable, and traceable to both technical and non-technical stakeholders. Transparency in AI promotes visibility into how models function, how data is processed, and how outcomes are derived, enabling accountability and trust across the system's lifecycle (Nikolinakos, 2023; Radanliev, 2025; Schmidt et al., 2020). According to the OECD AI Principles and the EU AI Act (Article 13), AI systems must be sufficiently transparent to enable users to interpret the system's output appropriately. The NIST AI Risk Management Framework similarly identifies transparency as a key property of trustworthy AI, emphasizing interpretability, traceability, and documentation. Likewise, ISO/IEC 42001 promotes transparency through structured documentation, model cards, and explainability records that support auditability and informed oversight. In essence, transparency serves as the connective tissue between AI developers, users, and regulators, ensuring that system logic, performance, and limitations are visible and communicated effectively. It forms the foundation for human oversight, risk management, and ethical assurance in AI-driven environments.

Transparency must be embedded throughout the AI lifecycle to ensure consistent visibility and interpretability at each phase:
- *Objective Setting*, through the definition of transparency goals, such as explainability, traceability, and disclosure obligations, as well as the identification of stakeholders who require visibility (e.g., developers, regulators, end users) and the specification of communication needs;
- *Requirements & Specifications*, through the inclusion of transparency requirements in functional and non-functional specifications (e.g., the need for model documentation, explainable AI (XAI) features, or audit trails), as well as the definition of how traceability will be maintained from data to decisions;
- *Design & Development*, through the incorporation of interpretable model architectures and documentation of the rationale for algorithmic choices, feature selection, and preprocessing steps, as well as the maintenance of data lineage and metadata to support reproducibility and accountability;
- *Verification & Validation*, through the conduct of explainability assessments to ensure model decisions can be justified, as well as the validation of documentation completeness, traceability matrices, and compliance with disclosure standards;
- *Deployment*, through the assurance that end-user interfaces present outputs in a comprehensible and non-deceptive manner, as well as the provision of users with information about the AI system's capabilities, limitations, and data usage; and
- *Operation & Maintenance*, through the continuous updating of transparency documentation as models evolve, as well as the maintenance of logs and audit records for retraining events, performance shifts, and decision rationales, and provision of traceability for any post-deployment modifications.

### 4.6. Privacy and Security

Privacy and Security form the foundational elements of trustworthy and compliant AI systems, ensuring that personal, sensitive, or proprietary data are collected, processed, and stored responsibly throughout the AI lifecycle. Privacy focuses on safeguarding individual data rights and ensuring compliance with data protection laws such as GDPR and HIPAA (Fabiano, 2025; Radanliev, 2025), while security emphasizes the protection of data and systems from unauthorized access, manipulation, or cyber threats (Alhitmi et al., 2024; Humphreys et al., 2024; Jada & Mayayise, 2024; Villegas-Ch & García-Ortiz, 2023). In the context of AI, privacy and security are critical not only for maintaining regulatory compliance but also for fostering public trust in algorithmic decision-making. The emergence of agentic AI systems further heightens these concerns, as autonomous decision loops can amplify data exposure risks or create new attack vectors if not properly governed.

Privacy and security considerations must be embedded across every phase of the AI lifecycle:
- *Objective Setting*, through the definition of data protection objectives, access control policies, and regulatory compliance requirements (e.g., GDPR, ISO/IEC 27001, NIST AI RMF), as well as the conduct of a Data Protection Impact Assessment (DPIA), where applicable;
- *Requirements & Specifications*, through the establishment of security- and privacy-by-design principles, encryption standards, and data minimization requirements, as well as the specification of auditability, traceability, and access management features;
- *Design & Development*, through the implementation of data anonymization, differential privacy, federated learning, and secure model development practices, as well as the integration of threat modeling techniques to identify potential vulnerabilities and misuse scenarios;
- *Verification & Validation*, through the performance of security testing (e.g., penetration testing, vulnerability scanning) and privacy compliance validation, as well as the validation of adherence to data governance controls and the assessment of resilience to data leakage or model inversion attacks;
- *Deployment*, through the enforcement of strong authentication, role-based access control, and network security measures, as well as the conduct of real-time monitoring for intrusion detection and anomaly identification; and
- *Operation & Maintenance*, through the continuous monitoring of privacy breaches and security vulnerabilities, as well as the implementation of incident response plans, retraining protocols, and regular patch management to maintain integrity and trustworthiness.

### 4.7. Data Governance

Data Governance encompasses the policies, processes, and accountability frameworks that ensure data is managed ethically, securely, and in compliance with regulatory and organizational requirements throughout the AI lifecycle. It ensures that data used to train, validate, and operate AI systems is accurate, consistent, traceable, and fit for its intended purpose (Corrêa et al., 2023; de Almeida & dos Santos, 2025; Law & McCall, 2024; Perry & Uuk, 2019; Tallberg et al., 2023; Wirtz et al., 2021). In the context of AI systems, effective data governance underpins trustworthiness, supporting fairness, privacy, and accountability. It mitigates risks such as data bias, data drift, and data misuse—all of which can directly affect model performance and regulatory compliance. The OECD AI Principles and the EU AI Act emphasize strong governance of data and data processes as essential to developing reliable and human-centric AI.

Data governance must be woven across each phase of the AI lifecycle, ensuring that every data-related activity is controlled and auditable:
- *Objective Setting*, through the definition of governance objectives aligned with organizational policies, ethical AI guidelines, and applicable regulations (e.g., GDPR, ISO/IEC 38505), as well as the establishment of data ownership, accountability roles, and governance committees;
- *Requirements & Specifications*, through the specification of data governance requirements such as metadata documentation, lineage tracking, consent management, and data quality standards, as well as the development of a data governance plan detailing lifecycle control mechanisms;
- *Design & Development*, through the implementation of procedures for data acquisition, labeling, and curation aligned with governance policies, as well as the assurance of data traceability and version control, and the application of checks for data representativeness and bias mitigation before model training;
- *Verification & Validation*, through the validation of data integrity, completeness, and provenance, as well as the verification of compliance with data retention and anonymization requirements, and the performance of governance audits to ensure adherence to approved data sources and usage policies;
- *Deployment*, through the establishment of monitoring pipelines to track ongoing data inputs, ensuring that operational data adheres to governance standards and that no unauthorized data sources are introduced; and
- *Operation & Maintenance*, through the continuous monitoring for data drift, data quality degradation, and unauthorized data reuse, as well as the conduct of periodic data governance reviews and the maintenance of auditable logs for traceability and accountability.

### 4.8. Fairness and Bias

Fairness and Bias reflect the ethical and technical imperative to ensure that AI systems provide equitable, non-discriminatory, and context-appropriate outcomes for all users and populations. Fairness in AI goes beyond simple accuracy parity—it encompasses the identification, mitigation, and continuous monitoring of biases that may arise from datasets, model architectures, workflows, or human–AI interactions (Barredo Arrieta et al., 2020; Mehrabi et al., 2021; NIST, 2023). As emphasized by the EU Ethics Guidelines for Trustworthy AI (EU, 2019) and the OECD AI Principles (OECD, 2021), fairness requires the prevention of both allocative harm (unequal distribution of opportunities or resources) and representational harm (reinforcement of stereotypes or exclusion of minority groups). Bias can be introduced through multiple vectors—historical biases embedded in training data, underrepresentation of subgroups, model design decisions, feedback loops, and operational context shifts. Therefore, fairness must be treated as a multi-dimensional risk factor requiring systemic controls, ongoing measurement, and periodic recalibration. The NIST AI RMF (2023) highlights the necessity of socio-technical evaluation, calling for domain-specific fairness metrics, transparent reporting, and context-appropriate definitions of equity.

Regarding its integration into the AI lifecycle, Fairness and Bias mitigation must be embedded continuously across every phase:

- *Objective Setting*, through the definition of fairness goals relevant to the intended use case, identification of sensitive attributes (e.g., age, gender, ethnicity), and assessment of potential differential impacts on subpopulations. This step also includes early ethical impact assessment and stakeholder consultation to identify equity-critical scenarios;
- *Requirements & Specifications*, through the establishment of explicit fairness requirements, selection of appropriate fairness metrics (e.g., equalized odds, demographic parity, predictive parity), and documentation of unacceptable bias thresholds. This phase also includes requirements for inclusive data sourcing and representativeness;
- *Design & Development*, through the integration of bias detection techniques, dataset balancing strategies, augmentation methods, and model-agnostic fairness interventions (e.g., reweighting, adversarial debiasing). Transparent model design choices, interpretability mechanisms, and subgroup performance reporting are essential outputs;
- *Verification & Validation*, through systematic fairness testing—including cross-group error analysis, sensitivity analysis, and scenario-based evaluations to capture hidden or emergent biases. Independent review, testing against baseline inequity thresholds, and validation against regulatory/ethical guidelines (e.g., GDPR, FDA Good Machine Learning Practice principles) are required to ensure fairness conformance;
- *Deployment*, through the implementation of human oversight checkpoints specifically designed to detect potential biased outputs in real workflows, especially in high-impact decision pathways. Deployment must also ensure traceability of fairness-related decisions and documentation of mitigation strategies, exceptions, or manual overrides; and

- *Operation & Maintenance*, through continuous fairness monitoring, automated alerts for subgroup performance drift, and periodic revalidation as demographic patterns or usage contexts evolve. Post-deployment audits, stakeholder feedback loops, and real-world impact assessments are essential for sustained fairness performance. Update cycles must incorporate fairness metrics alongside accuracy and safety metrics to avoid long-term systemic inequity.

### 4.9. Guardrails

Guardrails in AI systems refer to the technical, procedural, and ethical boundaries that constrain AI behavior to ensure it operates within safe, lawful, and intended limits. They serve as proactive mechanisms for risk prevention and mitigation, ensuring that AI systems remain aligned with human values, regulatory requirements, and organizational objectives (Berkey, 2024; Hakim et al., 2025; Pandya et al., 2025). Guardrails function both as design-time safeguards (e.g., data constraints, validation checks) and runtime controls (e.g., human-in-the-loop oversight, anomaly detection). They are especially critical in Agentic AI or autonomous systems, where dynamic behavior could lead to unpredictable or unsafe outcomes. By embedding guardrails, organizations establish fail-safe mechanisms that uphold trust, accountability, and ethical alignment, resonating with frameworks like the EU Ethics Guidelines for Trustworthy AI and the NIST AI Risk Management Framework, which emphasize "human oversight" and "risk management through governance controls."

Guardrails must be integrated as cross-cutting controls across the AI lifecycle—from concept to post-deployment—to maintain continuous oversight and system resilience:
- *Objective Setting*, through the definition of the ethical, safety, and compliance boundaries for the AI system, as well as the identification of unacceptable outcomes and the establishment of guardrail objectives consistent with organizational policies and legal frameworks (e.g., GDPR, FDA, ISO/IEC 23894);
- *Requirements & Specifications*, through the translation of ethical and regulatory expectations into functional and non-functional requirements, as well as the specification of acceptable performance thresholds, risk tolerance levels, and human intervention points;
- *Design & Development*, through the implementation of technical guardrails such as input validation, access control, model explainability, and safe response mechanisms, as well as the integration of policy-based constraints into training pipelines (e.g., excluding harmful or biased data) and the incorporation of oversight protocols like Human-in-Command (HIC), Human-in-the-Loop (HITL), and Human-on-the-Loop (HOTL) to ensure supervised autonomy;
- *Verification & Validation*, through the validation that guardrails perform as intended under normal and stress conditions, as well as the conduct of adversarial testing, scenario simulation, and edge-case evaluations to confirm robustness, and the documentation of results through guardrail verification reports;
- *Deployment*, through the implementation of runtime guardrails such as real-time monitoring, output filters, and fail-safe triggers that prevent the propagation of unsafe or noncompliant outputs, as well as the establishment of escalation mechanisms for human review of flagged events; and
- *Operation & Maintenance*, through the continuous monitoring of AI system behavior and adjustment of guardrails based on performance logs, incident trends, and emerging risks, as well as periodic review of thresholds and oversight mechanisms to align with updated policies, standards, or risk assessments.

## 5. Conclusion

The responsible development and deployment of AI systems in regulated domains such as healthcare requires a structured, lifecycle-oriented approach that embeds governance principles into every phase of system evolution. By operationalizing trustworthy AI pillars into actionable lifecycle, organizations can establish a measurable framework that moves beyond theoretical ethics commitments to demonstrable compliance evidence. In other words, SMART+ principles mapped across the lifecycle mandates early clarity of risk boundaries, safety-by-design modelling choices, independent validation before release, and ongoing surveillance for unintended outcomes. Collectively, these pillars form a governance scaffold that aligns with leading frameworks—including NIST AI RMF, ISO/IEC 42001, ISO/IEC 5338, EU AI Act, and ISPE GAMP 5 by translating abstract ethical expectations into practical validation controls. Importantly, they establish a dynamic feedback ecosystem linking design, deployment, and post-market learning, acknowledging that AI is not a static artifact but an adaptive system requiring ongoing oversight.

Additionally, AI systems are categorized into low, medium, and high-risk levels using the influence and decision-consequence (Kandikatla & Radeljić, 2025). Once the risk level is determined, the corresponding governance expectations are mapped to the SMART+ Framework. High-risk systems require the full application of all SMART+ items with the highest level of stringency, ensuring comprehensive controls across data, model, monitoring, and technical safeguards. Medium-risk systems mandate a moderate implementation of SMART+ controls, focusing on essential requirements while allowing context-based flexibility for advanced items. Low-risk systems require only a minimal or optional application of SMART+ elements, where the primary obligation is to conduct a basic risk assessment and apply lightweight controls as appropriate.

This tiered approach ensures that oversight scales proportionally with the potential impact and avoids unnecessary burden on low-impact systems while maintaining strong safeguards where needed most. Going forward, the integration of these pillars into both quality and AI management systems represents a significant step toward evidence-based AI governance.

Future research may explore checklists generated from this work to provide practitioners with a structured method to embed AI quality, facilitate audit readiness, and support continuous assurance. As AI systems scale across industries, such structured governance will be essential to uphold trust, safeguard stakeholders, and ensure that innovation advances without compromising human well-being.